\documentclass[preprint,12pt]{article}
\usepackage{authblk}

\usepackage{amssymb}

\usepackage{graphicx}
\usepackage{mathtools}
\usepackage[colorlinks, pdfencoding=auto, psdextra]{hyperref} 

\usepackage{adjustbox}

\usepackage{bm} 
\usepackage{bbm}
\usepackage{multirow}
\usepackage{acronym}
\usepackage{amsfonts}
\usepackage{booktabs}
\usepackage{caption}
\usepackage{subcaption}
\usepackage[normalem]{ulem}

\newcommand{\RR}{\mathbb{R}} 
\newcommand{\E}{\mathbb{E}} 
\newcommand{\N}{\mathcal{N}} 
\newcommand{\const}{\text{const}} 

\newcommand{\Phirff}{\text{\boldmath$\Phi$}_{\mbox{\rm \tiny{RFF}}}}
\newcommand{\PhirffT}{\text{\boldmath$\Phi$}_{\mbox{\rm \tiny{RFF}}}^\top}
\newcommand{\phirff}{\text{\boldmath$\phi$}_{\mbox{\rm \tiny{RFF}}}}

\DeclareMathOperator{\Tr}{Tr} 

\DeclarePairedDelimiter\pp{(}{)}
\DeclarePairedDelimiter\cor{[}{]}
\DeclarePairedDelimiter\llav{\{}{\}}
\DeclarePairedDelimiter\ang{\langle}{\rangle}
\DeclareMathOperator{\sumn}{\sum\limits_{n=1}^{N}} 
\DeclareMathOperator{\summ}{\sum\limits_{m=1}^{M}} 

\DeclareMathOperator{\prodn}{\prod\limits_{n=1}^{N}} 

\newcommand{\lnp}[1]{\ln\pp*{#1}} 
\newcommand{\eqline}{\enskip&\enskip}
\newcommand{\eqeq}{\enskip=&\enskip}

\newcommand{\eqsimil}{\enskip\sim &\enskip}

\DeclareMathOperator{\X}{\mathbf{X}} 
\DeclareMathOperator{\Ic}{\mathbf{I}_{C}} 

\DeclareMathOperator{\Xn}{\mathbf{x}_{n,:}} 
\DeclareMathOperator{\XnT}{\mathbf{x}_{n,:}^\top} 

\DeclareMathOperator{\Y}{\mathbf{Y}} 
\DeclareMathOperator{\YY}{\mathbf{Y}} 
\DeclareMathOperator{\Yn}{\mathbf{y}_{n,:}} 
\DeclareMathOperator{\Yc}{\mathbf{y}_{:,c}} 
\DeclareMathOperator{\Ync}{y_{n,c}} 
\DeclareMathOperator{\YnT}{\mathbf{y}_{n,:}^\top} 

\DeclareMathOperator{\K}{\text{\boldmath$\Phi$}} 

\DeclareMathOperator{\W}{\mathbf{W}} 
\DeclareMathOperator{\WW}{\mathbf{W}} 
\DeclareMathOperator{\Wf}{\mathbf{w}_{m,:}} 
\DeclareMathOperator{\Wc}{\mathbf{w}_{:,c}} 
\DeclareMathOperator{\Wfc}{w_{m,c}} 
\DeclareMathOperator{\WT}{\mathbf{W}^\top} 
\DeclareMathOperator{\WfT}{\mathbf{w}_{m,:}^\top} 
\DeclareMathOperator{\WcT}{\mathbf{w}_{:,c}^\top} 

\DeclareMathOperator{\bi}{\mathbf{b}} 
\DeclareMathOperator{\biT}{\mathbf{b}^\top} 
\DeclareMathOperator{\bic}{b_c} 

\DeclareMathOperator{\alp}{\text{\boldmath$\alpha$}} 
\DeclareMathOperator{\af}{\text{$\alpha_m$}} 

\DeclareMathOperator{\sumc}{\sum\limits_{c=1}^{C}} 
\DeclareMathOperator{\prodf}{\prod\limits_{m=1}^{M}} 
\DeclareMathOperator{\prodc}{\prod\limits_{c=1}^{C}} 

\acrodef{NN}{Neural Network}
\acrodef{NNs}{Neural Networks}
\acrodef{KMs}{Kernel Methods}
\acrodef{RFF}{Random Fourier Features}
\acrodef{RFF-BLR}{Random Fourier Features Bayesian Linear Regression}
\acrodef{SOTA}{state-of-the-art}
\acrodef{ARD}{Automatic Relevance Determination}
\acrodef{VI}{Variational Inference}
\acrodef{ELBO}{Evidence LowerBound}
\acrodef{MLP}{MultiLayer Perceptron}
\acrodef{SB-ELM}{Sparse Bayesian Extreme Learning Machine}
\acrodef{ELM}{Extreme Learning Machine}

\usepackage{xcolor}

\providecommand{\keywords}[1]{\textbf{\textit{Keywords -}} #1}

\title{Bayesian learning of feature spaces for multitask regression}

\author[1]{Carlos Sevilla-Salcedo
\thanks{Corresponding author

\textit{Email address}: carlos.sevillasalcedo@aalto.fi (Carlos Sevilla-Salcedo)}}
\author[2]{Ascensión Gallardo-Antolín}
\author[2]{Vanessa Gómez-Verdejo}
\author[2]{Emilio Parrado-Hernández}
\affil[1]{Department of Computer Science, Aalto University, Espoo, 02150, Helsinki, Finland}
\affil[2]{Department of Signal Theory and Communications, Universidad Carlos III de Madrid, Leganés, 28911, Madrid, Spain}

\begin{document}
\maketitle

\begin{abstract}

This paper introduces a novel approach for multi-task regression that connects Kernel Machines (KMs) and Extreme Learning Machines (ELMs) through the exploitation of the Random Fourier Features (RFFs) approximation of the RBF kernel. In this sense, one of the contributions of this paper shows that for the proposed models, the KM and the ELM formulations can be regarded as two sides of the same coin. These proposed models, termed RFF-BLR, stand on a Bayesian framework that simultaneously addresses two main design goals. On the one hand, it fits multitask regressors based on KMs endowed with RBF kernels. On the other hand, it enables the introduction of a common-across-tasks prior that promotes multioutput sparsity in the ELM view. This Bayesian approach facilitates the simultaneous consideration of both the KM and ELM perspectives enabling (i) the optimisation of the RBF kernel parameter $\gamma$ within a probabilistic framework, (ii) the optimisation of the model complexity, and (iii) an efficient transfer of knowledge across tasks. The experimental results show that this framework can lead to significant performance improvements compared to the state-of-the-art methods in multitask nonlinear regression.

\end{abstract}




\section{Introduction}
\label{sec:Introduction}
This paper proposes a method to construct simple non-linear models for multitask regression in scenarios that require interpretability but present small training datasets. The request for this type of models naturally arises in application domains subject to strong regulation, such as health or finances \cite{Ketu21,Xiong2014}.  In such domains, the ability to understand the outcome of machine learning models is a significant premise.  Another common characteristic is data scarcity due to the difficulty of setting up clinical studies, or of collecting data related to particular financial situations.
A common practice in these scenarios is the use of non-linear models formed by a single linear combination of simple non-linear functions of the original input features. Within the scope of \ac{NNs}, these models are implemented as a shallow architecture with a single hidden layer. Each neuron in the hidden layer encodes one of the non-linear functions that form the model. Examples of such architectures are the Radial Basis Function (RBF) \ac{NN}s \cite{Hartman90}, Two Layer Perceptrons (TLP) \cite{rumelhart1985} and \ac{ELM} \cite{huang2006,Huang2004}. Besides, broadly used \ac{KMs} such as the Support Vector Regressor (SVR) \cite{smola2004}, Kernel Ridge Regressor (KRR) \cite{vovk2013kernel} or Gaussian Process (GP) \cite{rasmussen2003gaussian} can also be considered \ac{NN}s with a single hidden layer \cite{burges1998tutorial}. 

Another peculiarity of these scenarios is the multitasking nature \cite{caruana1997}. Consider, for instance, the case of a clinical study focused on the characterisation of a disease. A common setup starts with a cohort of a few tenths (hundreds if we are lucky) of patients going through the same data acquisition process to collect a data set. Subsequently, these data can be used in the construction of models for the prediction of scores that can help capture patterns that characterise the progression of the disease, the probability of a successful response to treatment, etc. The interplay among those closely related tasks suggests that instead of learning each of these models within a separate, independent optimisation, a joint optimisation that covers all the tasks simultaneously could exploit these relationships among tasks, yielding more accurate models. In fact, several recent works \cite{spyromitros2016multi,zhen2017multi,zhen2018multi} show that the multitask regression paradigm can lead to significant improvements in performance achieved by learning an independent model per task. The outperformance of multitask regression is especially noticeable in cases with small training sets that demand non-linear modelling. These results have recently motivated a growing interest in multi-target regression algorithms, and in their successful application in numerous real-world problems. Among these applications, it is worth mentioning commerce and finance (e.g., prediction of stock price \cite{Li2019}), environmental modelling (e.g., air quality \cite{Masmoudi2020} and weather forecast \cite{Li2019}), robotics \cite{Li2019}, computer vision \cite{Emambakhsh2019,Farlessyost2021,Tan2022}, speech-related tasks (e.g., speech enhancement \cite{Tu2020} and speech intelligibility prediction \cite{Zezario2022}), and biomedical applications (e.g., medical image analysis \cite{Ma2022,Zhen2017}).

A common conclusion arises from these research works focusing on multitask regression with simple non-linear models and small data sets: the combination of the different tasks with a layer of common latent variables and the use of \ac{KMs} \cite{zhen2017multi,zhen2018multi} or \ac{ELM}s \cite{da2020outlier} to construct a non-linear mapping between the inputs and these latent variables yields more accurate models than other approaches such as ensembles of regression trees \cite{spyromitros2016multi}, or the use of gene expression programming to jointly learn a non-linear model for each task \cite{Moyano21}.

If we focus on these shallow neural networks, a critical stage during model design is the selection of the model architecture. Each method offers a different number of hyperparameters that translate into a different number of degrees of freedom or flexibility to choose the final architecture of the model. This set of hyperparameters includes at least the number of neurons in the hidden layer and the non-linear activation function implemented in each of these neurons of the hidden layer. In this sense, if we sort the shallow NN methods according to such flexibility to choose the final model architecture, perhaps KMs and ELMs would represent the two extremes of such range. In the common scenario of using Radial Basis Functions (RBFs) as non-linear functions, KMs offer minimal flexibility since one is only able to choose the spread parameter of the kernel. In these KMs the number of nodes in the hidden layer is either fixed (for instance, in Gaussian Processes this number is the size of the training set) or results from the final optimisation that fits the weights connecting the layers of neurons (like in the Support Vector Machines in which this number is the number of support vectors). And concerning the non-linear activation of each neuron, all of them are equipped with an RBF kernel centred on the corresponding training instance or support vector and with a shared spread parameter. Therefore designing a KM model primarily involves tuning the length scale of the RBF kernel through cross-validation. On the other extreme of the flexibility spectrum, ELMs offer the largest flexibility in choosing the model architecture. Users have the freedom to select the number of neurons in the hidden layer, the specific non-linear activation implemented at each neuron, and the parameters or centroids that define these non-linear functions. The hidden layer of an ELM can accommodate neurons equipped with different non-linearities  \cite{huang2015}, such as sigmoid functions, RBFs, trigonometric functions, polynomials, wavelets, Fourier functions, and hard-limit functions. Additionally, the concept of hidden neurons can be extended to architectures where the nodes in the hidden layers are sub-networks \cite{huang2015local}. As the size of the training set grows toward situations that would invite crossing the shallow-to-deep learning threshold, ELMs endow shallow architectures with those extra degrees of flexibility able to capture more complex input/output relationships. 

This large flexibility in the design of ELMs acts as a double-edged weapon. It enables the estimation of regression functions of arbitrary complexity (ELMs possess universal approximation capabilities\cite{huang2015local}) but at the expense of increasing the potential risk of overfitting. In this context, the incorporation of elastic net penalties into the ELM optimisation, as discussed in \cite{martinez2011}, serves to alleviate overfitting by controlling the hidden layer's size. Moreover, ELMs combined with structured regularisation techniques such as group lasso \cite{yuan2006model}, have been applied to multitask scenarios \cite{inaba2018,da2020outlier}. Nevertheless, these approaches typically involve a large number of hyperparameters to be tuned, often through a resource-intensive cross-validation procedure. Furthermore, establishing a meaningful connection with domain-specific prior knowledge can prove challenging in certain types of scenarios.

Another means of alleviating overfitting within the ELM family is the Sparse Bayesian ELM \cite{luo2013sparse}. This Bayesian formulation enables the introduction of prior knowledge in architecture design \cite{Soria2011}, with an optimisation that controls overfitting by increasing the sparsity in the output layer. This sparsity serves as a regularisation by nullifying the effect of irrelevant nodes in the hidden layer. This way, the network prediction will in fact be a function of a smaller subset of neurons in the hidden layer. However, to the best of our knowledge, the Bayesian ELM and its sparse version are only formulated for single-target regression. An extension to efficiently cover the multitask case demands a prior that forces joint sparsity over the different sets of parameters related to each task, resulting in all tasks sharing the same hidden layer (with a different output layer per task), leading to an effective transfer of knowledge among tasks. This idea underlies multitask regression models such as group lasso \cite{yuan2006model}, dirty models \cite{jalali2010dirty}, or multilevel lasso \cite{lozano2012multi}.

Our proposal, termed \ac{RFF-BLR}, leverages the connection between Kernel Machines (KMs) and Extreme Learning Machines (ELMs) by approximating the RBF kernel using \ac{RFF}s \cite{rahimi2007random}. We observe that an RBF neural net can be viewed as a KM endowed with an RBF kernel, and therefore be approximated by an ELM whose hidden layer contains the \ac{RFF}s that approximate the kernel.

Besides, this paper proposes a Bayesian framework that jointly addresses two key aspects. On the one hand, it enables the optimisation of simple multitask regressors based on KMs endowed with RBF kernels. On the other hand, it introduces a shared-among-tasks prior that encourages multioutput sparsity in the ELM view of the model. In other words, the Bayesian approach enables the consideration of both the KM and RFF views of the model. Consequently, it enables the tuning of the RBF kernel parameter $\gamma$ in a probabilistic framework, avoiding the need for cross-validations. Moreover, the value of $\gamma$ can capture valuable knowledge to draw an initial coarse-resolution model architecture in the KM view. Simultaneously, the sparsity induced in the output layer further refines this initial architecture, ultimately yielding a final regression model of suitable complexity.

The main contributions of the paper can be summarized as follows:
\begin{itemize}
    \item A formulation that establishes the connection between a KM endowed with an RBF kernel and an equivalent ELM by recognizing their link through the  \ac{RFF} approximation of the kernel.
    \item A model for multi-target regression where a single, efficient non-linear mapping is jointly learned across all tasks via Bayesian optimisation.
    \item A probabilistic framework for automatic learning the RBF kernel lengthscale without the necessity of computationally expensive cross-validation procedures. 
    \item An empirical evaluation of the advantages of this proposal in several small sample multitask regression benchmarks.
\end{itemize}

\section{Related work}
\label{sec:Related}
Besides the works already mentioned in the introduction, there are several approaches in the literature that deal with multitask problems \cite{borchani2015survey}.
The most common methods are based on linear models that exploit the correlation between the output tasks by means of regularisation. This is the case of the aforementioned group lasso model \cite{yuan2006model}, dirty models \cite{jalali2010dirty}, multilevel lasso \cite{lozano2012multi}, the Multiple output Regression with Output and Task Structures (MROTS) \cite{Rai2012}, Multivariate Regression with Covariance Estimation (MRCE) \cite{rothman2010sparse} that jointly learns the model weights and the task correlations, or the work of \cite{liu2015calibrated} where additional regularisation parameters are included to deal independently with the noise in each task. 

Other methods rely on the building of ensemble architectures to improve multi-output regression performance, such as the FItted Rule Ensembles (FIREs) \cite{aho2012multi}, the multi-objective random forests \cite{kocev2007ensembles}, or the work of \cite{Moyano21} which combines the ensemble architecture with gene expression programming. 

Conversely, other works propose to simplify the modelling of the output space relationships employing a cluster structure; this way, each task cluster is independently modelled, and the tasks within the same cluster possess similar weight vectors. Here, Clustered Multi-Target Learning (CMTL) \cite{jacob2008clustered} and Flexible Clustered Multi-Target (FCMTL) \cite{zhou2015flexible} stand out, where the cluster structure is learnt by identifying representative tasks. 

However, the kernel-based multitask methods, capable of exploiting non-linear relationships between data and/or between tasks, have shown the best performance. This is the case for algorithms such as \cite{brouard2016input} where the authors use kernels to learn the non-linear structure of the observations or \cite{dinuzzo2011learning} where the kernel is used in the output. 
Within this framework, the Multi-layer Multi-target Regression (MMR) method delivers very competitive results in terms of accuracy \cite{zhen2017multi,zhen2018multi}.
In these works the data are mapped into a \sout{high} reproducing kernel Hilbert space where a common feature representation and inter-target correlation are learnt.

Another set of techniques that can be used for multitasking problems are models based on deep learning. Under this paradigm, non-linear relationships can be easily modelled and can be efficiently applied to areas involving image or time series processing \cite{ruder2017overview}. However, they do not meet any of the requirements of the scenarios we want to work with, as deep learning models demand a large amount of data to avoid overfitting, require long training time, and yield networks that are difficult to interpret. 

\section{Connection between Kernel Methods and Extreme Learning Machines through Random Fourier Features}
\label{sec:kernel-ELM}
This section illustrates the connection between \ac{KMs} and ELMs in a single target regression framework for simplicity. The next section will describe its extension to the multitask regression framework, which is the main focus of this research.

Let us consider a training data set formed by $N$ observations with their corresponding targets $\{(\Xn, y_n)\}_{n=1}^N$, where $\Xn$ are vectors with $D$ components and $y_n \in \RR$. Conversely, let us call $\X$ the $N\times D$ matrix whose rows contain the training observations.  A KM would construct a regression model that follows the expression 
\begin{equation}\label{eq-km}
f(\mathbf x) = \sum_{n=1}^{N} \beta_n\kappa(\Xn, \mathbf x) + b,
\end{equation}
where $\{\kappa(\Xn, \cdot)\}_{n=1}^N$ are kernel functions centred on the training observations and the coefficients $\{\beta_n\}_{n=1}^N$ and $b$ are determined by an optimisation that tries to accurately approximate the targets without overfitting. This optimisation usually involves minimising a loss function evaluated in the training set, $\mathcal L(f(\mathbf x), y)$ and a regularisation term $\Omega(\beta_1,\dots,\beta_N)$ that helps prevent overfitting:
\begin{equation}\label{eq-optim-km}
\min_{\beta_1,\dots,\beta_N,b} \sum_{n=1}^{N}{\mathcal L(f(\Xn), y_n)} + \lambda \Omega(\beta_1,\dots,\beta_N)
\end{equation}
Sparsity can be achieved by imposing combinations of losses and regularisations that force that a fraction of the $\beta_n$s become zero after the optimisation. Once the kernel function is selected, controlling the expressive power of the model reduces to choosing a suitable value for the kernel (and for the loss and regulariser) hyperparameters. For instance, in the ubiquitous RBF kernel case 
\begin{equation}\label{eq-krbf}
\kappa(\mathbf x_{i,:}, \mathbf x_{j,:})= \exp{\left(-\gamma\|\mathbf x_{i,:} - \mathbf x_{j,:}\|^2\right)}
\end{equation}
the $\gamma$ hyperparameter controls the smoothness of $f(\mathbf x)$, the smaller the value of $\gamma$, the smoother the model. The value of $\gamma$ is commonly determined by cross-validation, although in the Gaussian Process case, the kernel parameters are optimised by maximising the marginal likelihood, exploiting their probabilistic framework.

A model with the structure of eq. \eqref{eq-km} and the kernel of eq. \eqref{eq-krbf} can be implemented as an RBF \ac{NN} in which training observations act as centroids for neurons in the hidden layer and coefficients $\{\beta_n\}_{n=1}^N$ as weights that connect the hidden layer with the output layer.

ELMs significantly enlarge the degrees of flexibility in the design of the single hidden layer NN architecture, as one is free to select the number of neurons in the hidden layer and the non-linearity implemented in each of these neurons. 

\ac{RFF}s establish a connection between \ac{KMs} and ELMs: a KM endowed with an RBF kernel can be approximated by an ELM whose nodes implement sinusoidal non-linearities. The kernel trick \cite{Scholkopf2002} establishes that the choice of a kernel $\kappa(\cdot, \cdot)$ induces the selection of a lifting $\boldsymbol h(\cdot)$ of the input data in a feature space. This way, the evaluation of the kernel between two observations in the input space is equivalent to the computation of a dot product in the feature space between the two lifted observations:
\begin{equation}
\kappa(\mathbf x_{i,:}, \mathbf x_{j,:}) = \langle \boldsymbol h(\mathbf x_{i,:}), \boldsymbol h(\mathbf x_{j,:}) \rangle.
\end{equation}

The \ac{RFF}s approximate the lifting $\boldsymbol h(\cdot)$ with a lower-dimensional mapping $\phirff(\mathbf x)$ (note that for some kernels the corresponding $\boldsymbol h(\cdot)$ involves lifting into a feature space of infinite dimensions) so that 
\begin{equation}\label{eq-approx-kernel}
\kappa(\mathbf x_{i,:}, \mathbf x_{j,:}) = \langle \boldsymbol h(\mathbf x_{i,:}), \boldsymbol h(\mathbf x_{j,:}) \rangle \approx \phirff(\mathbf x_{i,:})^\top \phirff(\mathbf x_{j,:}).
\end{equation}
Each of the $M$ coordinates of the mapping $\phirff(\mathbf x)$ (named RFF) is a sinusoid with a frequency randomly sampled from the Fourier transform of the kernel. Since the Fourier transform of an RBF kernel with its hyperparameter $\gamma$ is given by
\begin{equation}
p(\omega) = \sqrt{\frac{\pi}{\gamma}} \exp{\left (\frac{-\pi^2\omega^2}{\gamma}\right )}
\end{equation} the RFFs that approximate $\boldsymbol h(\cdot)$ are sinusoids with $D$-dimensional frequencies $\{\boldsymbol \omega_m \}_{m=1}^M$ sampled from the input space according to a Gaussian distribution with zero mean and spherical covariance with variances proportional to $\gamma$ \cite{rahimi2007random}. In other words, the $m$-th coordinate of mapping $\phirff(\mathbf x)$ is:
\begin{equation}\label{eq-phi-k}
\phi_m(\mathbf x) = \cos(\boldsymbol \omega_m^\top \mathbf x + b_m), \quad m=1,\dots,M
\end{equation}where $b_m$ is randomly sampled from a uniform in $[0,2\pi]$. Concerning the size of $M$, in the case of the RBF kernel the larger the better, but since the convergence of the approximation is exponential with $M$ \cite{rahimi2007random}, empirical results show that $M>2N$ yields reasonably good approximations. 

The model in eq. \eqref{eq-km} can be written in vector form
\begin{equation}\label{eq-km-vector}
f(\mathbf x) = \boldsymbol \kappa(\mathbf x)^\top \boldsymbol \beta + b.
\end{equation}where $\boldsymbol \kappa(\mathbf x)$ is the vector whose components are the evaluation of the kernel between $\mathbf x$ and the $N$ training observations, and $\boldsymbol \beta$ is a vector containing coefficients $\{\beta_n\}_{n=1}^N$. Now let us denote $\Phirff$ as the  $N\times M$ matrix whose rows are $\{\phirff(\Xn)\}_{n=1}^N$. This way $\boldsymbol \kappa(\mathbf x)$ in eq. \eqref{eq-km-vector} can be approximated by
\begin{equation}\label{eq-approx-kernel-vector}
\boldsymbol \kappa(\mathbf x) \approx \Phirff \phirff(\mathbf x)
\end{equation}
and after merging eqs. \eqref{eq-km-vector} and \eqref{eq-approx-kernel-vector}, $f(\mathbf x)$ results in 
\begin{equation}\label{eq-primal-rff}
f(\mathbf x) \approx \phirff(\mathbf x)^\top \Phirff^\top \boldsymbol \beta + b = \phirff(\mathbf x)^\top \mathbf w + b.
\end{equation} Therefore, model $f(\mathbf x)$ can be also implemented by an ELM with a hidden layer formed by $M$ neurons equipped with the non-linearities of eq. \eqref{eq-phi-k} and with weights $\mathbf w = \Phirff^\top \boldsymbol \beta $ connecting that hidden layer with the output layer.

Alternatively, this could have been started with the design of model $f(\mathbf x)$ from the ELM perspective. Let us assume that we instantiate an ELM with $M^\prime$ nodes in the hidden layer, each endowed with the corresponding non-linearity of eq. \eqref{eq-phi-k}. Fitting this model would involve an optimisation to find $\mathbf w$ and $b$ in eq. \eqref{eq-primal-rff}. If we adapt problem \eqref{eq-optim-km} for this purpose, we will end up optimising:
\begin{equation}\label{eq-optim-elm}
\min_{\mathbf w,b} = \sum_{n=1}^{N}{\mathcal L(f(\Xn), y_n)} + \lambda \Omega(\Phirff\Phirff^\top)^{-1}\Phirff\mathbf w)
\end{equation} where $f(\Xn)$ in $\mathcal L(f(\Xn), y_n)$ is calculated with the right part of eq \eqref{eq-primal-rff}.

Then, as long as $M^\prime$ is large enough to produce a good approximation of the lifting, the  $\mathbf w$ resulting from \eqref{eq-optim-elm} will be a close approximation to the $\Phirff^\top \boldsymbol \beta$ that we would obtain if instead we optimize the KM view with \eqref{eq-km}.


\section{Random Fourier Features Bayesian Linear Regression (RFF-BLR) for multitask problems}
\label{sec:RFF-BLR}

To cover the multitask regression case, let us extend the notation of the paper as follows: In a data set with ${\rm C}$ simultaneous tasks, the target $\mathbf y_{n,:}$  corresponding to observation $\Xn$ , ${\rm n}=1, \ldots, {N}$ becomes a vector with ${\rm C}$ components, one per task. Therefore, consider $\bf Y$ the $N\times C$ output matrix whose rows are ${\bf y}_{n,:}$ for $n=1, \ldots, {\rm N}$.

The remainder of this section presents the \ac{RFF-BLR} model, which combines \ac{RFF}s with Bayesian Linear Regression for the automatic selection of the frequency components of the neurons that define the hidden layer of the final model. The combination of \ac{RFF} and BLR brings various benefits: 
\begin{itemize}
    \item Including the parameter $\gamma$ in the formulation somewhat shrinks the search space for the frequencies that define the sinusoidal non-linear activations of the neurons of the hidden layer of the ELM.
    \item The sparsity-inducing Bayesian formulation nullifies useless neurons in the hidden layer of the ELM, further refining the model. This also allows us to gain in interpretability and alleviates the possible overfitting of the ELM architecture.
    \item Moreover, the Bayesian formulation enables cross-task sparsity, as it favours models that share \ac{RFF}s across tasks in a robust manner.
\end{itemize}

The derivation starts with the definition of the corresponding generative model, followed by the development of the variational inference of the model parameters. 
\begin{figure}[ht]
  \centering
    \includegraphics[page=1,width=0.7\columnwidth]{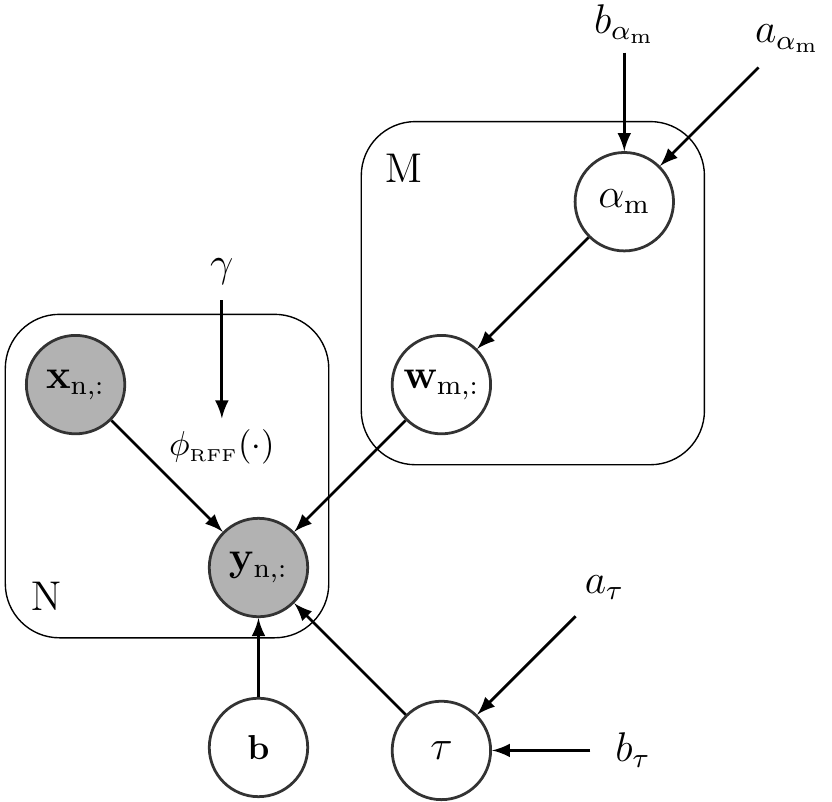}\\
    \caption{Plate diagram for the \ac{RFF-BLR} model. Grey circles denote observed variables, white circles unobserved random variables. Symbols without a circle correspond to the hyperparameters.}
    \label{fig:GM_LR-RFF}
\end{figure}

Bayesian linear regression in the space defined by the \ac{RFF}s establishes that the observed $\Y$ was generated by a linear combination of the \ac{RFF}, $\Phirff$, with a weight matrix $\W \in \RR^{\rm M \times C}$ plus a bias term and some additive noise as
\begin{align}
    \Y \eqeq \Phirff\W + \bi \mathbbm{1}_N+ \eta,
\end{align}
where $\eta$ represents Gaussian noise with precision $\tau$, $\bi \in \RR^{\rm 1 \times C}$ is the bias of the model and $\mathbbm{1}_N$ is a row vector of ones with dimension $\rm N$. Figure \ref{fig:GM_LR-RFF} depicts this generative model. Distributions associated with the model can be defined as
\begin{align}
    \Yn \eqsimil \N \pp*{\phirff(\Xn) \W + \bi,\tau^{-1} \Ic}\\
    \Wf \eqsimil \N \pp*{\mathbf{0},\af^{-1}\Ic}\\
    \af \eqsimil \Gamma\pp*{\alpha_0, \beta_0}\\
    \bi \eqsimil \N \pp*{\mathbf{0},\Ic}\\
    \tau \eqsimil \Gamma\pp*{\alpha^{\tau}_0, \beta^{\tau}_0}
\end{align}
where $\Ic$ is an identity matrix of dimension $\rm C$; $\alpha_0$, $\beta_0$ are $\af$ prior hyperparameters, and $\alpha^{\tau}_0$, $\beta^{\tau}_0$ are $\tau$ prior hyperparameters.

After defining the generative model, we can evaluate the posterior distribution of all the model variables using an approximate inference approach through mean-field variational inference \cite{Blei17}. This involves the maximisation of a lower bound for the posterior distribution and the approximation of this posterior with a fully factorised variational family as
\begin{align}
&p(\Theta\mid\Phirff, \Y) \approx  q\pp*{\W} q\pp*{\tau} q\pp*{\bi} \prodf q\pp*{\af}, \label{eq:qModel}
\end{align}
where $\Theta = [\W, \alp, \bi ,\tau]$ comprises all the random variables in the model.

The mean-field posterior structure along with the lower bound 
results in a feasible coordinate-ascent-like optimisation algorithm in which the optimal maximisation of each of the factors in eq. \eqref{eq:qModel} can be computed if the rest remain fixed using the following expression
\begin{align}\label{eq:meanfield}
q^*(\theta_i) \propto \mathbb{E}_{\Theta_{-i}}\left[\log p(\Theta,\mathbf{y}_{1,:},\dots, \mathbf{y}_{\rm N,:}\mid\Phirff)\right],
\end{align}
where $\Theta_{-i}$ comprises all random variables but $\theta_i$. This new formulation is generally feasible as it does not require a complete marginalisation of $\Theta$ from the joint distribution, which is calculated as
\begin{align}
p(&\YY, \WW, \alp, \tau, \bi \mid \Phirff) \nonumber \\
\eqeq \prodn p\pp*{\Yn \mid \phirff(\Xn) , \W, \bi, \tau} p\pp*{\W \mid \alp} p\pp*{\alp} p\pp*{\bi} p\pp*{\tau}. \label{eq:joint}
\end{align}

Therefore, eq. \eqref{eq:joint} can be substituted in eq. \eqref{eq:meanfield}
for each random variable to obtain the model update rules. The resulting approximate distributions are included in Table \ref{tab:RFFBLRupdate}. As the estimated distribution for each r.v. also depends on some other r.v., e.g. $\ang{\Wc}$ depends on $\ang{\tau}$ and $\ang{\bic}$, we need to iterate over the variables, analysing the evolution of the lower bound on each iteration until convergence. The complete development of all $q^*$ distributions as well as the final mean-field factor update rules are included in \ref{sec:VIU}.


\begin{table}[hbt]
\centering
\caption{Updated distributions $q$ for the r.v. of the graphical model. Here, $diag(\mathbf{x})$ is an operator that transforms a vector into a diagonal matrix with diagonal $\mathbf{x}$, $\mathbbm{1}_N$ is a row vector of ones of dimension $\rm N$, and $<>$ represents the mean value of the r.v. These expressions have been obtained using the update rules of the mean-field approximation in eq. \eqref{eq:meanfield}. 
}
\begin{adjustbox}{max width=\textwidth}
\begin{tabular}{ccc}
\toprule
{\textbf{Variable}} & {$\bm{q}^*$ \textbf{distribution}} & \textbf{Parameters} \\\midrule
\multirow{3}{*}{$\W$} 
& \multirow{3}{*}{$\prodc \N \pp*{\Wc \mid \ang{\Wc}, \Sigma_{\Wc}}$} 
& $\ang{\Wc} = \ang{\tau} \Sigma_{\W} \Phirff^{\top} \pp*{\Yc - \mathbbm{1}_{\rm N}\ang{\bic}}$ \\
& & $\Sigma_{\W}^{-1} = diag(\ang{\alp}) + \ang{\tau}\Phirff^{\top} \Phirff$  \\&&\\
\midrule    
\multirow{3}{*}{$\af$}
& \multirow{3}{*}{$\Gamma\pp*{\af\mid \mathbf{a}_{\af}, \mathbf{b}_{\af}}$}
& $\mathbf{a}_{\af} = \alpha_0 + \frac{\rm C}{2}$ \\ 
& & $\mathbf{b}_{\af} = \beta_0 + \frac{1}{2} \ang{\Wf^{\top}\Wf}$ \\&&\\
\midrule
\multirow{3}{*}{$\bi$}
& \multirow{3}{*}{$\N \pp*{\bi \mid \ang{\bi}, \Sigma_{\bi}}$}
& $\ang{\bi} = \ang{\tau} \sumn\pp*{\Yn - \phirff(\Xn)\ang{\W}} \Sigma_{\bi}$ \\ 
& & $\Sigma_{\bi}^{-1} = \pp*{{\rm N} \ang{\tau} + 1} \Ic$ \\&&\\
\midrule
\multirow{5}{*}{{$\tau$}}
& \multirow{5}{*}{$\Gamma\pp*{\tau\mid \mathbf{a}_{\tau}, \mathbf{b}_{\tau}}$}
& $\mathbf{a}_{\tau} = \alpha_0^\tau + \frac{\rm NC}{2}$ \\
& & $\mathbf{b}_{\tau} = \beta_0^\tau + \frac{1}{2} \sumn\sumc \Ync^2 + \frac{1}{2}{\rm Tr}\llav{\ang{\WT\W} \Phirff\Phirff^{\top}}$ \\
& & $ - {\rm Tr}\llav{\Y \ang{\WT} \Phirff^{\top}} - \sumn \Yn \ang{\biT}$  \\
& & $ + \sumn \phirff(\Xn) \ang{\W} \ang{\biT} + \frac{\rm N}{2} \ang{\bi \biT}$  \\
\bottomrule
\end{tabular}
\end{adjustbox}
\label{tab:RFFBLRupdate}
\end{table}

\subsection{Automatic selection of relevant RFF components}

In the definition of the generative model, r.v.
 $\Wf$ and $\af$ form an Automatic Relevance Determination (ARD) prior that promotes sparsity over the rows of the input matrix, that is, the \ac{RFF} or neurons of the hidden layer of the ELM. 
A component $\alpha_m$ of $\alp$ ending up with a high value after the optimisation implies that the corresponding RFF turned out to be irrelevant, thus all elements in $\Wf$ will be zero. This removes the effect of that component \ac{RFF} for all tasks. In summary, the model allows for the automatic selection of the most relevant RFF jointly for all tasks, thus allowing a transfer of knowledge between them.

\subsection{Optimisation of the kernel parameter \texorpdfstring{$\gamma$}{gamma}  in RFF-BLR}

The Bayesian nature of the model can be exploited to perform an automatic optimisation of the parameter $\gamma$ by maximising the lower bound of the defined mean-field approach. The calculation of the lower bound is included in \ref{sec:LB}. As the lower bound needs to be optimised with respect to $\Phirff$, the terms that do not depend on $\Phirff$ can be considered constant,resulting in
\begin{align}
    LB' \eqeq  \ang{\tau} \sumn\sumc \left( \Ync\ang{\WcT} \phirff(\Xn)^{\top} \right. \nonumber \\
    \eqline-\frac{1}{2}\ang{\Wc\WcT} \phirff(\Xn)^{\top}\phirff(\Xn) 
    \nonumber\\
    \eqline - \left.  \phirff(\Xn) \ang{\Wc} \ang{\bic} \right) + \const. \label{eq:LBderivative}
\end{align}
where $\const$ include the terms not depending on $\Phirff$. Then, the model can alternate between mean-field updates over the variational bound and direct maximisation of eq. \eqref{eq:LBderivative} w.r.t. $\gamma$ using any gradient ascend method (e.g. Adam \cite{Kingma14}). Therefore, the value of $\gamma$ can be determined without the need for any type of cross-validation.


\section{Experimental work}
\label{sec:Results}
This section includes results on broadly used multitask regression benchmarks to help gain insight into the true capabilities of the proposal in real-world scenarios. 

\subsection{Baselines}
\label{sec:Baselines}
The first baseline is the \textbf{Multi-layer Multi-target Regression} (MMR) method proposed in \cite{zhen2018multi}. This method uses an RBF kernel to construct a non-linear mapping from the input space into a set of latent intermediate features. These latent features are connected with the output space with a linear mapping that captures the interdependencies among the different tasks. The fitting of this model depends on three hyperparameters that control the regularisation and the kernel lengthscale. Following \cite{zhen2018multi} we adjusted the values of the regularisation parameters exploring 9 values in the range of $10^{-5}$ to $10^3$ in the logarithmic scale and used the average of all pairwise distances between training instances to estimate the kernel lenghtscale.

The second baseline is the \textbf{Generalised Outlier Robust ELM} (GOR-ELM) of \cite{da2020outlier}. A feed-forward neural network that combines $\ell_{2,1}$ regularisation with Elastic Net theory to work with multitask problems. The neurons forming its hidden layer are equipped with sigmoid functions as nonlinearities. Following \cite{da2020outlier} we used 1000 neurons in the hidden layer and cross-validated the value of the model hyperparameters $\alpha$ in the range $\{0, 0.25, 0.5, 0.75, 1\}$ and $\lambda$ in the range $2^{-20:1:20}$.

In addition, we have considered
another set of baselines that directly learn a linear combination of the RFFs 
and impose sparsity on the weight matrix ($\W$):
\begin{itemize}
    \item \textbf{Multi-Task Feature Learning (MTFL)} \cite{argyriou2006multi} uses the group LASSO to regularise the features used by different tasks, imposing sparsity on the weight matrix.
    We cross-validated the regularisation hyperparameter exploring a grid of $10^{-5:1:5}$ and for the RBF kernel hyperparameter $\gamma$ we used a grid of $\frac{2^{-10:1:0}}{D}$.
    \item \textbf{\ac{SB-ELM}} \cite{luo2013sparse} induces sparsity in the output layer by using an ARD prior to perform automatic feature selection. One model is needed for each output task in a one-vs-all fashion.
\end{itemize}
Finally, we also included two NN 
models using the RBF nonlinearities in several hidden layers that offer a larger expressive power than the single hidden layer NNs of this study:
\begin{itemize}
    \item \textbf{Feed-forward Neural Network (FNN)} \cite{murtagh1991multilayer} with non-linear relations between the inputs and the multiple outputs. We validated five configurations: (i) one hidden layer with 100 neurons, (ii) two hidden layers with 100 and 50 neurons, (iii) three hidden layers with 100, 50 and 100 neurons, (iv) four hidden layers with 100, 50, 50 and 100 neurons, and (v) five hidden layers with 100, 50, 25, 50 and 100 neurons.
    \item \textbf{Heterogeneous Incomplete - Variational AutoEncoder (HI-VAE)} \cite{nazabal2020handling} is an adaptation of the Variational AutoEnconder that captures the latent representation of the data while being able to work with heterogeneous data. We used the layer configuration suggested in \cite{nazabal2020handling}: three layers of dimensions 50-50-50, respectively.
\end{itemize} 

\subsection{Datasets}

We used eight multi-output regression datasets from the \textit{Mulan} repository \cite{spyromitros2016multi,karalivc1997first,dvzeroski2000predicting}. Their main characteristics are summarised in Table \ref{tab:MTR-datasets}. 
\begin{table}[htp]
\caption{Characteristics of the multitask databases from the \textit{Mulan} repository.}
\label{tab:MTR-datasets}
\centering
\begin{adjustbox}{max width=\columnwidth}
\begin{tabular}{cccc}
\hline
Database & Samples & Features & Tasks \\ \hline
\textit{at1pd}        & 337        & 411        & 6          \\
\textit{at7pd}        & 296        & 411        & 6          \\
\textit{oes97}        & 334        & 263       
& 16         \\
\textit{oes10}        & 403        & 298        & 16         \\
\textit{edm}          & 154        & 16         & 2          \\
\textit{jura}         & 359        & 15         & 3          \\
\textit{wq}           & 1,060      & 16         & 14         \\
\textit{enb}          & 768        & 8          & 2          \\
\hline
\end{tabular}
\end{adjustbox}
\end{table}

Each dataset was evaluated following a 10-fold Cross-Validation (CV). For models that need to cross-validate hyperparameters, we adopted a nested cross-validation scheme within each training partition of the main 10-fold cross-validation. 
We standardised the input data for all models except for MMR which validates whether to standardise the data and GOR-ELM, which uses min-max scaling, as suggested by the authors. We used the coefficient of determination ($R^2$) to compare the performance of the different methods and adjust their hyperparameters. This accuracy score achieves a maximum value of 1 when the model can approximate all the targets in the test set with no errors. Therefore, the higher the value of this score, the more accurate the model. For each data set, we report as the final score the arithmetic average (and standard deviations) of all $R^2$ scores obtained in all tasks and CV folds.


\subsection{Results}
\label{sec:multregression}
Table \ref{tab:MultiTaskv0} displays the results of the empirical comparison between the proposed model, \ac{RFF-BLR}, and the baselines. 
These results show that \ac{RFF-BLR} outperforms the baselines in all databases except one. In particular, it achieves an improvement of around $0.14$ in \textit{oes97} and $0.21$ in \textit{jura} over the $R^2$ achieved by 
the best baselines. For the rest of the databases,
it consistently provides good performance, obtaining a net average improvement of $0.11$ with respect to the best baselines, GOR-ELM and MTFL. 

If we just look at the baseline performance, the first remark is the clear superiority of the multitask models over the single-task SB-ELM. Moreover, two of the single hidden layer NNs with multitask formulation, MTFL and GOR-ELM, outperform multitask NNs with several hidden layers (HI-VAE and FNN). Our intuition behind this fact is that the small size of the training data sets hampers the models with more expressive power. 
With respect to the influence of the non-linear function on the performance of the model, the behaviour is not consistent along all the datasets. For example, the baseline equipped with a sigmoid as non-linearity (GOR-ELM) achieves the best performance in \textit{atp1d} and \textit{jura} datasets, whereas the MMR endowed with an RBF kernel becomes the best baseline in data set \textit{wq}, and the MTFL with RFF produces the best results in \textit{oes97} and \textit{oes10} datasets. In this context, it is worth noting the fact that \ac{RFF-BLR} incorporates both RBF and RFF dual views of the same model in a formulation that does a good job adjusting the expressive power of the model to the needs of each dataset, hence achieving this significantly better performance.


\begin{table}[!th]
    \begin{center}
    \caption{Results in multitask benchmark data sets. The values represent the mean and standard deviation of the 10 $R^2$ scores obtained by each method for all tasks in each dataset and all CV folds. The text in parenthesis specifies the non-linearity used, namely, RBF kernels, sigmoid function or RFF projection.}
    \label{tab:MultiTaskv0}
    \adjustbox{max width=\textwidth}{
        \setlength\tabcolsep{2.5pt} 
        \begin{tabular}{lcccccccccc}
            \toprule
            Model & HI-VAE & FNN & MTFL & MMR & GOR-ELM & \ac{SB-ELM} & \ac{RFF-BLR} \\
            Kernel & (RBF) & (RBF) & (\ac{RFF}) & (RBF) & (sigmoid) & (\ac{RFF})& (\ac{RFF})\\
            \midrule
            \textit{atp1d}
            & $0.72 \pm 0.07$ & $0.80 \pm 0.10$ & $0.78 \pm 0.09$ & $0.80 \pm 0.09$ & $0.81 \pm 0.09$ & $0.60 \pm0.16$ &$\mathbf{0.83\pm 0.06}$ \\
            \textit{atp7d}
            & $0.42 \pm 0.12$ & $0.64 \pm 0.12$ & $0.55 \pm 0.13$ & $0.53 \pm 0.51$ & $0.63 \pm 0.14$ & $0.60 \pm 0.14$ & $\mathbf{0.74 \pm 0.15}$ \\
            \textit{oes97}
            & $0.50 \pm 0.22$ & $0.60 \pm 0.16$ & $0.69 \pm 0.12$ & $0.45 \pm 0.26$ & $0.68 \pm 0.14$ & $0.31 \pm 0.32$ & $\mathbf{0.83 \pm 0.05}$ \\
            \textit{oes10}
            & $0.66 \pm 0.10$ & $0.77 \pm 0.09$ & $\mathbf{0.83 \pm 0.07}$ & $0.57 \pm 0.31$ & $0.82 \pm 0.05$ & $0.46 \pm 0.23$ & $\mathbf{0.83 \pm 0.12}$ \\
            \textit{edm}
            & $0.34 \pm 0.11$ &  $0.10 \pm 0.34$ & $0.36 \pm 0.15$ & $0.36 \pm 0.22$ & $0.26 \pm 0.30$ & $0.38 \pm 0.20$ & $\mathbf{0.49\pm 0.14}$ \\
            \textit{jura}
            & $0.54 \pm 0.07$ & $0.35 \pm 0.19$ & $0.61 \pm 0.10$ & $0.60 \pm 0.10$ & $0.64 \pm 0.11$ & $0.64 \pm 0.09$ & $\mathbf{0.85 \pm 0.07}$ \\
            \textit{wq}
            & $0.07 \pm 0.02$ & $0.13 \pm 0.03$ & $0.12 \pm 0.01$ & $0.15 \pm 0.01$ & $0.12 \pm 0.03$ & $-0.02\pm 0.07$ & $\mathbf{0.21\pm 0.02}$ \\
            \textit{enb}
            & $0.91 \pm 0.01$ & $\mathbf{0.99 \pm 0.02}$ & $0.98 \pm 0.01$ & $0.91 \pm 0.05$ & $0.98 \pm 0.01$ & $0.97\pm 0.01$ & $0.96\pm 0.02$ \\
            \midrule
            average & $0.52 \pm 0.09$ & $0.55 \pm 0.13$ &  $0.62 \pm 0.08$ & $0.55 \pm 0.19$ & $0.62 \pm 0.11$ & $0.49 \pm 0.15$ & $\mathbf{0.73 \pm 0.05}$
        \end{tabular}
    }
    \end{center}
\end{table}

\begin{figure}[!ht]
    \centering
    \begin{subfigure}[b]{0.31\textwidth}
         \centering
         \includegraphics[width=\textwidth]{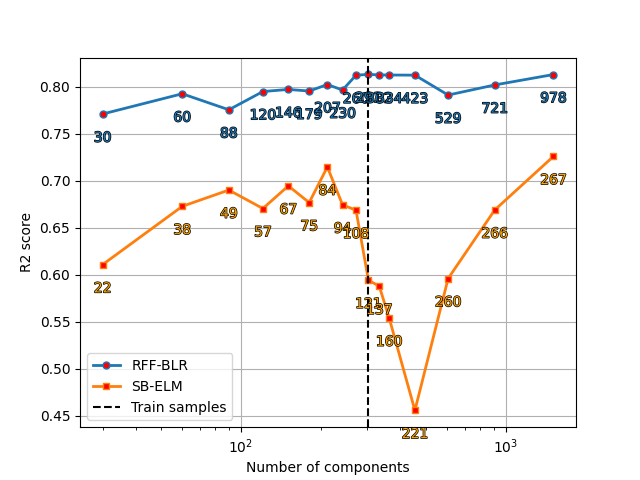}
         \caption{\textit{atp1d} database.}
         \label{fig:Comp_atp1d}
     \end{subfigure}
    \begin{subfigure}[b]{0.31\textwidth}
         \centering
         \includegraphics[width=\textwidth]{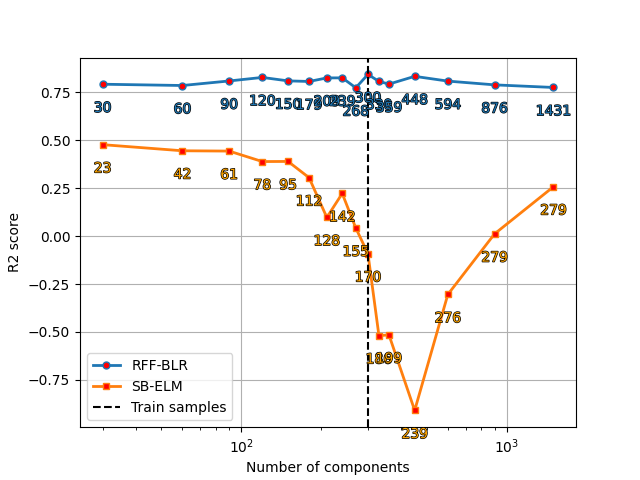}
         \caption{\textit{oes97} database.}
         \label{fig:Comp_oes97}
     \end{subfigure}
    \begin{subfigure}[b]{0.31\textwidth}
         \centering
         \includegraphics[width=\textwidth]{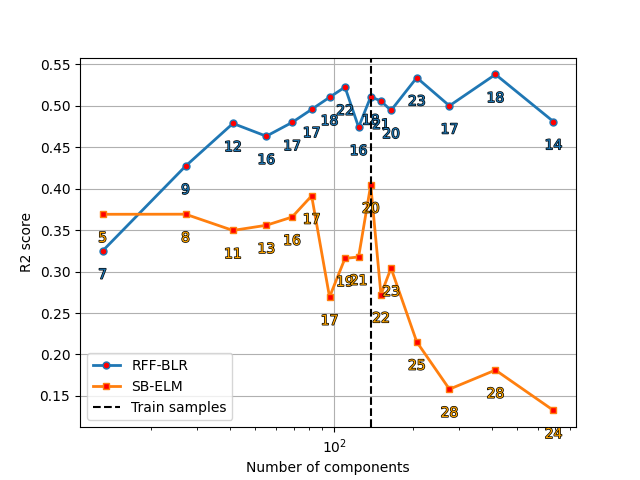}
         \caption{\textit{edm} database.}
         \label{fig:Comp_edm}
     \end{subfigure}
    \begin{subfigure}[b]{0.31\textwidth}
         \centering
         \includegraphics[width=\textwidth]{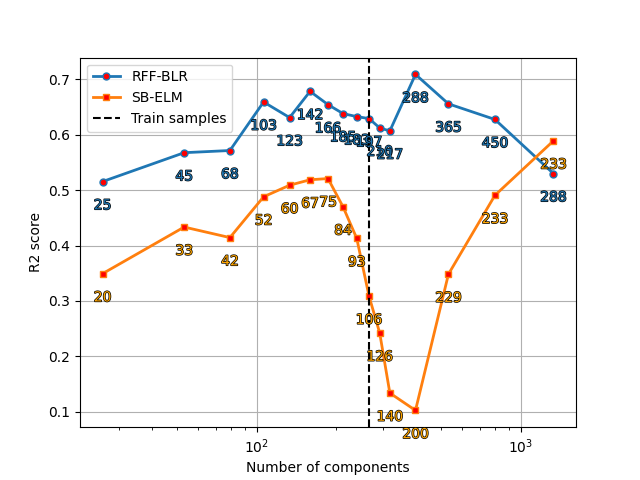}
         \caption{\textit{atp7d} database.}
         \label{fig:Comp_atp7d}
     \end{subfigure}
    \begin{subfigure}[b]{0.31\textwidth}
         \centering
         \includegraphics[width=\textwidth]{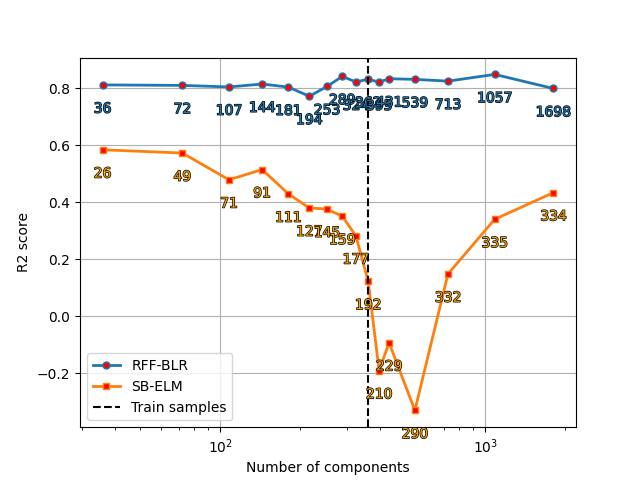}
         \caption{\textit{oes10} database.}
         \label{fig:Comp_oes10}
     \end{subfigure}
    \begin{subfigure}[b]{0.31\textwidth}
         \centering
         \includegraphics[width=\textwidth]{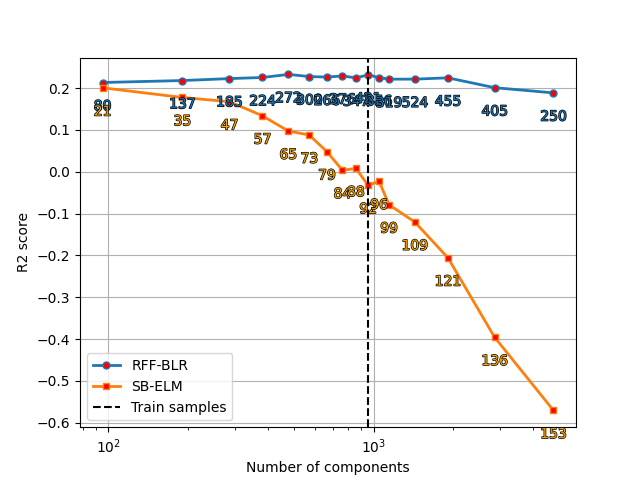}
         \caption{\textit{wq} database.}
         \label{fig:Comp_wq}
     \end{subfigure}
    \begin{subfigure}[b]{0.31\textwidth}
         \centering
         \includegraphics[width=\textwidth]{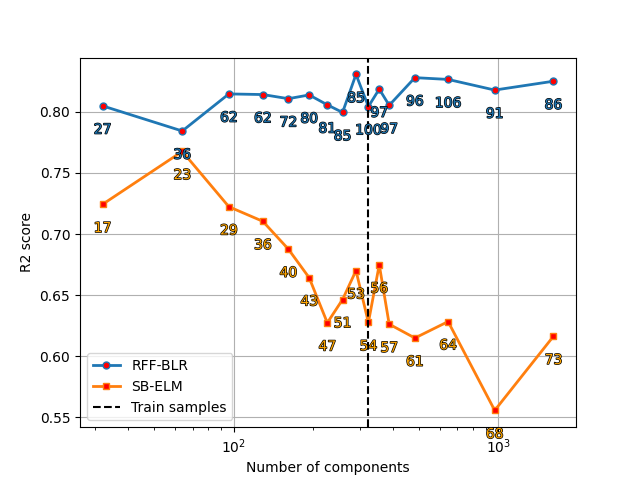}
         \caption{\textit{jura} database.}
         \label{fig:Comp_jura}
     \end{subfigure}
    \begin{subfigure}[b]{0.31\textwidth}
         \centering
         \includegraphics[width=\textwidth]{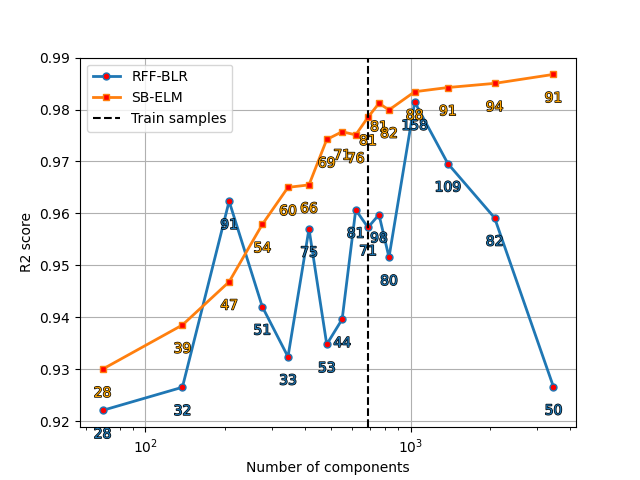}
         \caption{\textit{enb} database.}
         \label{fig:Comp_enb}
     \end{subfigure}
\caption{Evaluation of the $R^2$ score as a function of the initial and final number of RFFs. This experiment compares the performance of \ac{SB-ELM} (orange) and \ac{RFF-BLR} (blue) averaged over 10-folds. The vertical dotted line shows the point where the initial number of \ac{RFF}s is equal to the size of the training set. The numbers close to each point represent the final number of \ac{RFF} selected after pruning. 
}
\label{fig:Components}
\end{figure}

Let us now focus the discussion on the level of sparsity achieved by the \ac{RFF-BLR} method in the benchmarks. Figure \ref{fig:Components} shows the dependence of the final accuracy of the model with the initial value of $M$ (number of RFFs). 
The blue curves in the plots show the $R^2$ averaged for all tasks in each data set achieved by \ac{RFF-BLR}. The numbers close to the curve indicate the number of RFFs that define the final model once the optimisation is finished, and the vertical dashed line marks the size of the training set. We include as a baseline for comparison the SB-ELM (orange curves), as it also follows a sparse Bayesian framework specially targeted to develop ELMs with a very compact hidden layer. First, in 7 out of the 8 problems \ac{RFF-BLR} achieves clearly higher $R^2$ scores than SB-ELM. Notice how in most cases RFF-BLR shows a performance, in terms of $R^2$, not very sensitive to the initial value of $M$,  unlike for SB-ELM. Moreover, the final numbers of RFFs show a great sparsity in most cases compared to that of the SB-ELM. 

The last part of the discussion is devoted to a more in-depth assessment of the computational efficiency of our approach. To achieve this, we conducted a synthetic experiment to measure the computational time required for training our method while varying the number of output tasks, exploring 15 values ranging from 2 to 16, as well as the number of input samples, incorporating 50 values between 50 and 1,000. We carried out these experiments five times with different initializations and calculated the average results across these repetitions, considering all possible combinations of parameters.

\begin{figure}[!ht]
    \centering
    \begin{subfigure}[b]{0.49\textwidth}
         \centering
         \includegraphics[width=\textwidth]{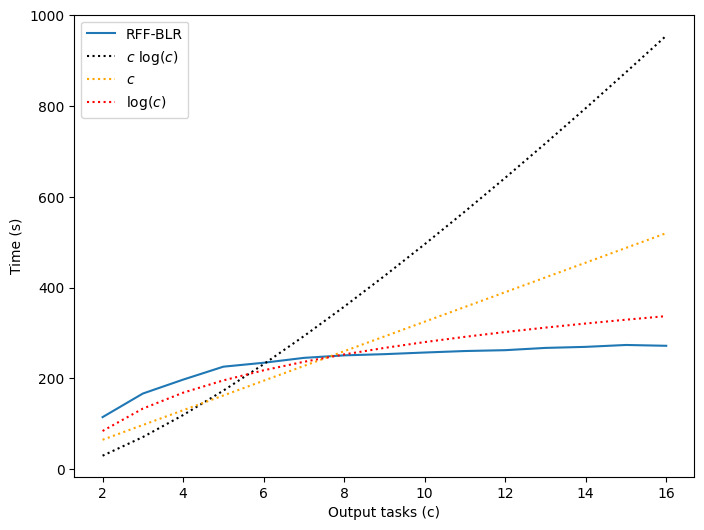}
         \caption{Analysis of the number of output tasks.}
         \label{fig:compCost_C}
     \end{subfigure}
    \begin{subfigure}[b]{0.49\textwidth}
         \centering
         \includegraphics[width=\textwidth]{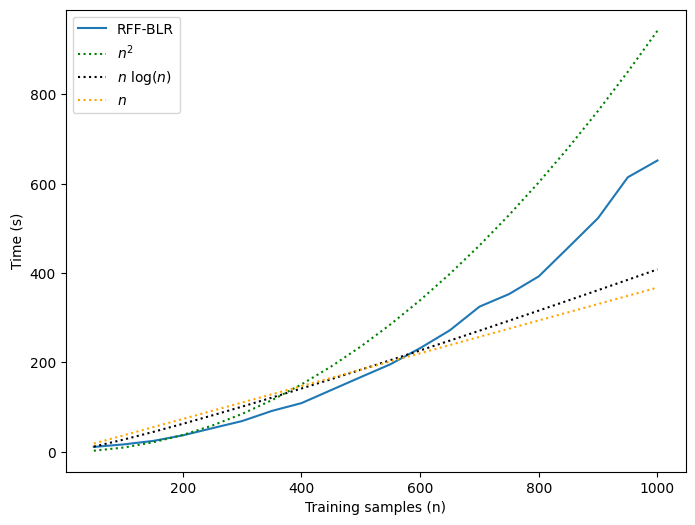}
         \caption{Analysis of the number of training samples.}
         \label{fig:compCost_N}
     \end{subfigure}
\caption{Evaluation of the computational cost of the model on a synthetic dataset. For the number of tasks ($c$), we explored 15 values between 2 and 16. For the number of training samples ($n$), we explored 50 values between 50 and 1,000. The results have been averaged over 5 repetitions and the different values explored for the other parameter.}
\label{fig:ComputationalCost}
\end{figure}

Figure \ref{fig:compCost_C} shows the relationship between the computational cost and the number of output tasks. The plots show that the computational cost of our approach (as a function of the number of tasks) is slightly lower than that of a logarithmic curve. This behaviour means that our proposal is particularly efficient and robust when dealing with large numbers of output tasks.

Concerning the dependence of the computational cost with the size of the training set, Figure \ref{fig:compCost_N} shows that the computational cost of our proposal is lower than that of a quadratic curve. This feature is particularly interesting because it indicates that our model not only effectively optimises the number of RFFs through gradient descent, but also outperforms common kernel methods that typically exhibit a quadratic increase in computational cost with the size of the training set.

\section{Conclusions}
\label{sec:Conc}




This paper has presented the RFF-BLR, an algorithm designed to learn accurate sparse multitask regression models. RFF-BLR leverages the observation that a neural network (NN) with a single hidden layer containing N neurons equipped with RBF activation functions with the same parameter $\gamma$ can be sharply approximated by a NN with a hidden layer whose neurons implement the RFFs that approximate that RBF kernel. RFF-BLR unifies these two perspectives within a single Bayesian framework, effectively extending Sparse Bayesian ELM to multitask regression scenarios. This formulation controls the expressive power of the ELMs by incorporating domain knowledge through the kernel parameter and by the use of a sparsity-inducing prior.


One the one hand, the experimental results show that the RFF-BLR achieves significantly better performance than the state-of-the-art single hidden layer NN models equipped with different non-linearities in several broadly used multitask regression benchmark problems with small training datasets. Moreover, this architecture also outperforms NNs with more hidden layers.  
On the other hand, RFF-BLR is capable of delivering very sparse multitask architectures, in which the same hidden layer formed by a very compact subset of these RFFs serves the regression models that cover each of the tasks.


\appendix
\section{RFF-BLR variational inference updates}
\label{sec:VIU}
This section includes the complete mathematical derivation of the update rules of the model parameters using variational inference. For simplicity, we can start by determining the log-probability of the output data given the model parameters
\begin{align} 
&\ln p\pp*{\Y \mid \Phirff ,\W,\tau,\bi} \nonumber\\
\eqeq \sumn \lnp{\N\pp*{ \phirff(\Xn)\W + \bi, \pp*{\tau}^{-1}\Ic}} \nonumber \\
\eqeq  \sumn\left(\frac{1}{2} \ln\mid\pp*{\tau}^{-1}I\mid-\frac{\tau}{2}\pp*{ \Yn - \phirff(\Xn)\W - \bi}\right.\nonumber\\
\eqline\left.  \pp*{\Yn - \phirff(\Xn)\W - \bi}^T \right) + \const \nonumber \\ 
\eqeq \frac{\rm N C}{2} \ln\pp*{\tau} - \frac{\tau}{2} \sumn \left(\Yn\YnT - 2\Yn \WT \phirff(\XnT)+ \bi\biT \right.\nonumber\\
\eqline\left.- 2\Yn\biT + 2\phirff(\Xn)\W \biT \right.\nonumber\\
\eqline\left.+ \phirff(\Xn)\W \WT \phirff(\XnT)\right) + \const \label{eq:logpY}
\end{align} 
where $\const$ include the terms without r.v.

\subsection{Distribution of \texorpdfstring{$\W$}{\textbf{W}}}
The approximate log probability of variable $\W$ is given by
\begin{align} 
\lnp{q^*\pp*{\W}} \eqeq \E_{\tau, \alp \bi}\cor*{\lnp{p\pp*{\Y, \W, \alp, \tau, \bi \mid \Phirff}}} \nonumber \\
\eqeq \E_{\tau, \bi}\cor*{\lnp{p\pp*{\Y \mid \Phirff,\W,\tau,\bi}}} \nonumber\\
\eqline + \E_{\alp}\cor*{\lnp{p\pp*{\W \mid \alp}}} + \const,
\end{align} 
where the first term is
\begin{align} 
&\lnp{p\pp*{\Y \mid \Phirff,\W,\tau,\bi}} \nonumber \\
\eqeq - \frac{\tau}{2} \sumn \sumc \left(- 2\Ync \WcT \phirff(\XnT)+ 2 \WcT \phirff(\XnT)\bic \right.\nonumber\\
\eqline\left.+ \WcT \phirff(\XnT)\phirff(\Xn)\Wc \right) + \const\nonumber \\
\eqeq \tau \sumc \left(\WcT \PhirffT \Yc + \WcT \Phirff^\top \mathbbm{1}_{\rm N}\bic \right.\nonumber\\
\eqline\left. + \frac{1}{2}\WcT \PhirffT \Phirff\Wc \right) + \const,
\label{eq:E1Wb}
\end{align} 
where $\mathbbm{1}_N$ is a row vector of ones of dimension $\rm N$. Then, the second term is
\begin{align} 
\lnp{p\pp*{\W \mid \alp}} \eqeq \summ \lnp{p\pp*{\Wf \mid \af}} \nonumber\\
\eqeq \summ \sumc \frac{\af}{2} {\Wfc^2} + \const \nonumber \\
\eqeq \frac{1}{2} \sumc \WcT diag(\alp)+ \const.
\label{eq:lnpWa}
\end{align} 

Then, by calculating the expectation, we get
\begin{align} 
\lnp{q^*\pp*{\W}} \eqeq \sumc  \left( \ang{\tau}\WcT \PhirffT( \Yc - \mathbbm{1}_{\rm N}\ang{\bic}) \right.\nonumber\\
\eqline\left. - \frac{1}{2}\WcT (diag(\alp) \right.\nonumber\\
\eqline\left.+  \ang{\tau}\PhirffT \Phirff) \Wc \right) + \const. 
\end{align} 

Identifying terms, we see that the $q$ distribution of the variable is
\begin{align} 
q^*\pp*{\W} \eqeq \prodc \N \pp*{\Wc \mid \ang{\Wc}, \Sigma_{\Wc}},
\end{align} 
where the variance is common for all output tasks, $\rm C$, and can be expressed as
\begin{align} 
\Sigma_{\W}^{-1} \eqeq diag(\ang{\alp}) + \ang{\tau}\PhirffT \Phirff,
\end{align} 
and mean
\begin{align} 
\ang{\W} \eqeq \ang{\tau} \Sigma_{\W} \PhirffT \pp*{\Y - \mathbbm{1}_{\rm N}\ang{\bi}}.
\end{align}
where $\ang{\W}$ is a stacked version of $\ang{\Wc}$.

\subsection{Distribution of \texorpdfstring{$\alp$}{\textbf{alpha}}}
The approximate distribution of $\alp$ follows
\begin{align} 
\lnp{q^*\pp*{\alp}} \eqeq \E_{\W}\cor*{\lnp{p\pp*{\Y, \W, \alp, \tau, \bi \mid \Phirff}}} \nonumber \\
\eqeq \E_{\alp}\cor*{\lnp{p\pp*{\W \mid \alp}}} \nonumber \\
\eqline + \E\cor*{\lnp{p\pp*{\alp}}} + \const,
\end{align} 
where the first term corresponds to Equation \eqref{eq:lnpWa} and the second term is
\begin{align} 
&\E \cor*{\lnp{p\pp*{\alp}}} 
= \summ\pp*{ \lnp{p\pp*{\af}} } \nonumber\\
\eqeq \summ\pp*{-\beta_0 \af + \pp*{\alpha_0 -1}\lnp{\af}} + \const
\end{align} 

Then, joining both terms, we get
\begin{align} 
\lnp{q^*\pp*{\alp}}
\eqeq  \summ\left(\pp*{\frac{C}{2} + \alpha_0 -1}\lnp{\af}\right. \nonumber \\
\eqline \left.  -\pp*{\beta_0 + \frac{1}{2} \ang{\WfT \Wf }}\af\right) + \const
\end{align} 

Therefore, the $q$ distribution of $\alp$ is
\begin{align}
q\pp*{\alp} \eqeq \prodf \Gamma\pp*{\af\mid \mathbf{a}_{\af}, \mathbf{b}_{\af}}
\end{align}
with the distribution parameters calculated as
\begin{align}
\mathbf{a}_{\af} \eqeq \alpha_0 + \frac{\rm C}{2} \\
\mathbf{b}_{\af} \eqeq \beta_0 + \frac{1}{2} \ang{\WfT\Wf}
\end{align}

\subsection{Distribution of \texorpdfstring{$\bi$}{\textbf{b}}}
The distribution of variable $\bi$ is given by
\begin{align} 
\lnp{q^*\pp*{\bi}} \eqeq \E_{\W,\tau}\cor*{\lnp{p\pp*{\Y, \W, \alp, \tau, \bi \mid \Phirff}}} \nonumber \\
\eqeq \E_{\W,\tau}\cor*{\lnp{p\pp*{\Y \mid \Phirff,\W,\tau,\bi}}} \nonumber\\
\eqline + \E\cor*{\lnp{p\pp*{\bi}}} + \const,
\end{align} 
where the effect of the prior of the bias is given by
\begin{align} 
\lnp{p\pp*{\bi}} \eqeq \lnp{\N\pp*{0,I}} = - \frac{1}{2}\bi \biT +\const, \nonumber
\end{align} 
and the remaining term of the distribution can be calculated similarly to Equation \eqref{eq:E1Wb}. Then, by calculating the expectation, we get
\begin{align} 
\lnp{q^*\pp*{\bi}} \eqeq \sumn  \left( \ang{\tau}(\Yn - \phirff(\Xn)\ang{\W})\biT \right.\nonumber\\
\eqline\left.- \frac{1}{2}\bi (\Ic + {\rm N} \ang{\tau}) \biT \right) + \const. 
\end{align} 

Once this expectation is calculated, we can determine that the distribution followed by the parameter is given by
\begin{align} 
q^*\pp*{\bi} \eqeq \N \pp*{\bi \mid \ang{\bi}, \Sigma_{\bi}},
\end{align} 
where the variance is
\begin{align} 
\Sigma_{\bi}^{-1} \eqeq \pp*{{\rm N} \ang{\tau} + 1} \Ic,
\end{align} 
and the mean is
\begin{align} 
\ang{\bi} \eqeq \ang{\tau} \sumn\pp*{\Yn - \phirff(\Xn)\ang{\W}} \Sigma_{\bi}.
\end{align} 

\subsection{Distribution of \texorpdfstring{$\tau$}{tau}}
Finally, the approximate distribution of $\tau$ is
\begin{align}
\lnp{q^*\pp*{\tau}} \eqeq \E_{\W,\bi}\cor*{\lnp{p\pp*{\Y \mid \Phirff,\W,\tau,\bi}}} \nonumber\\
\eqline+ \E\cor*{\lnp{p\pp*{\tau}}} + \const.
\end{align} 

We can calculate the expectation of Equation \eqref{eq:logpY}, obtaining
\begin{align} 
&\E_{\W,\bi}\cor*{\lnp{p\pp*{\Y \mid \Phirff,\W,\tau,\bi}}}
= \frac{\rm N C}{2} \lnp{\tau} \nonumber\\
\eqline - \frac{\tau}{2} \left(\sumn\sumc \Ync^2 - 2 \Tr\llav*{\ang{\WT}\PhirffT \Y} \right. \nonumber \\
\eqline \left.  + \Tr\llav*{\ang{ \W\WT} \PhirffT \Phirff} - 2 \sumn \Yn \ang{\biT} \right. \nonumber \\
\eqline \left. + 2 \sumn \phirff(\Xn)\ang{\W} \ang{\biT} + N \ang{\bi \biT} \right) ,
\end{align} 
and then the second term is
\begin{align} 
\E\cor*{\lnp{p\pp*{\tau}}} \eqeq \lnp{p\pp*{\tau}} = - \beta_0^{\tau} \tau + \pp*{\alpha_0^{\tau} -1}\lnp{\tau}+ \const.
\end{align} 

So, if we join both expectation elements and identify distribution terms, we see that the new distribution is
\begin{align} 
q^*\pp*{\tau} \eqeq \Gamma\pp*{\tau \mid a_{\tau},b_{\tau}},
\end{align} 
where the parameter $a_{\tau}$ is
\begin{align} 
a_{\tau} \eqeq \frac{\rm N C}{2} + \alpha_0^\tau,
\label{eq:aTauBias}
\end{align} 
and the parameter $b_{\tau}$ can be expressed as
\begin{align}
\mathbf{b}_{\tau} \eqeq \beta_0^\tau + \frac{1}{2} \sumn\sumc \Ync^2 + \frac{1}{2}{\rm Tr}\llav{\ang{\WT\W} \K\PhirffT} \nonumber\\
\eqline - {\rm Tr}\llav{\Y \ang{\WT} \PhirffT} - \sumn \Yn \ang{\biT} \nonumber \\
\eqline + \sumn \phirff(\Xn)\ang{\W} \ang{\biT} + \frac{\rm N}{2} \ang{\bi \biT} \label{eq:bTauBias}
\end{align}

\section{RFF-BLR lower bound}
\label{sec:LB}
Here, we present the complete derivation of the lower bound of the model. We can calculate the changes in the lower bound as follows:
\begin{align} 
LB \eqeq - \int q\pp*{\Theta}\lnp{\frac{q\pp*{\Theta}}{p\pp*{\Y,\Theta\mid\Phirff}}} d\Theta \nonumber \\
\eqeq \int q\pp*{\Theta}\lnp{p\pp*{\Y,\Theta\mid\Phirff}} d\Theta - \int q\pp*{\Theta}\lnp{q\pp*{\Theta}} d\Theta \nonumber \\
\eqeq \E_{q}\cor*{\lnp{p\pp*{\Y,\Theta\mid\Phirff}}} - \E_{q}\cor*{\lnp{q\pp*{\Theta}}} \label{eq:Lq}
\end{align} 

We will separately calculate the terms related to $\E_{q}\cor*{\lnp{p\pp*{\Y,\Theta\mid\Phirff}}}$ and the entropy of $q\pp*{\Theta}$.

\subsection{Terms associated to \texorpdfstring{$\E_{q}\cor*{\lnp{p\pp*{\Y,\Theta\mid\Phirff}}}$}{E[Y,Theta/K]}}
This first term of the lower bound would be composed by the following terms:
\begin{align} 
&\E_{q}\cor*{\lnp{p\pp*{\Y,\Theta \mid \Phirff}}} = \nonumber \\
\eqline \E_{q}\cor*{\lnp{p\pp*{\W\mid\alp}}} +\E_{q}\cor*{\lnp{p\pp*{\alp}}}  \nonumber \\
\eqline +\E_{q}\cor*{\lnp{p\pp*{\Y\mid\W,\Phirff, \bi,\tau}}} +\E_{q}\cor*{\lnp{p\pp*{\tau}}}  +\E_{q}\cor*{\lnp{p\pp*{\bi}}}
\label{eq:ElogpXTheta}
\end{align}


These are calculated as
\begin{align} 
&\E_{q}\cor*{\lnp{p\pp*{\W\mid\alp}}} = -\frac{\rm M C}{2}\lnp{2\pi} - \summ\pp*{a_{\af}} \nonumber \\
\eqline +\frac{\rm C}{2}\summ \pp*{\psi\pp*{a_{\af}}  - \lnp{b_{\af}}} + \beta_0 \summ\pp*{\frac{a_{\af}}{b_{\af}}}
\label{eq:ElogpWgivenAlpha}
\end{align} 

\begin{align} 
&\E_{q}\cor*{\lnp{p\pp*{\alp}}} 
= {\rm C} \pp*{\alpha_0 \lnp{\beta_0} - \lnp{\Gamma\pp*{\alpha_0}}} \nonumber \\
\eqline+ \summ\pp*{- \beta_0 \frac{a_{\af}}{b_{\af}} + \pp*{\alpha_0-1} \Big(\psi\pp*{a_{\af}} - \lnp{b_{\af}}\Big)}
\label{eq:ElogpAlpha}
\end{align} 

\begin{align} 
&\E_{q}\cor*{\lnp{p\pp*{\W,\alp}}} 
= \pp*{\frac{\rm C}{2}+\alpha_0-1}\summ \Big(\psi\pp*{a_{\af}} - \lnp{b_{\af}\Big)}  \nonumber\\
\eqline -\frac{\rm M C}{2}\lnp{2\pi} +  {\rm C} \Big(\alpha_0 \lnp{\beta_0} - \lnp{\Gamma\pp*{\alpha_0}}\Big) - \summ\pp*{a_{\af}}
\end{align} 

\begin{align} 
&\E_{q}\cor*{\lnp{p\pp*{\Y\mid\W,\Phirff,\bi,\tau}}} \nonumber \\
\eqeq -\frac{N C}{2}\lnp{2\pi}+\frac{C}{2}\pp*{\E_{q}\cor*{\lnp{\tau}}} \nonumber\\
\eqline - \frac{\ang{\tau}}{2} \sumn\sumc \Big(\Ync\Ync + \Ync\ang{\WcT} \phirff(\Xn)^{\top}\nonumber \\
\eqline -\frac{1}{2}\ang{\Wc\WcT} \phirff(\Xn)^{\top}\phirff(\Xn) - \Ync \ang{\bic} \nonumber \\
\eqline  - \phirff(\Xn)\ang{\Wc} \ang{\bic}+ \frac{1}{2} \ang{\bic \bic} \Big)
\label{eq:ElogpXgivenTau}
\end{align} 

\begin{align} 
&\E_{q}\cor*{\lnp{p\pp*{\tau}}}
= \alpha_0^{\tau}\lnp{\beta_0^{\tau}} - \lnp{\Gamma\pp*{\alpha_0^{\tau}}} - \beta_0^{\tau}\frac{a_{\tau}}{b_{\tau}} \nonumber \\
\eqline+ \pp*{\alpha_0^{\tau}-1}\Big(\psi\pp*{a_{\tau}} - \lnp{b_{\tau}}\Big)
\label{eq:ElogpTau}
\end{align} 


\begin{align}
\E_q\cor*{\lnp{p\pp*{\bi}}} \eqeq - \frac{\rm C}{2} \lnp{2\pi} - \frac{1}{2} \ang{\bi \biT}
\end{align}

\subsection{Terms of entropy of \texorpdfstring{$q\pp*{\Theta}$}{q(Theta)}}
The second term of the lower bound, the entropy, can be calculated as
\begin{align} 
\E_{q}\cor*{\lnp{q\pp*{\Theta}}} \eqeq \E_{q}\cor*{\lnp{q\pp*{\W}}} +\E_{q}\cor*{\lnp{q\pp*{\alp}}}\nonumber\\
\eqeq + \E_{q}\cor*{\lnp{q\pp*{\tau}}} + \E_{q}\cor*{\lnp{q\pp*{\bi}}}, \label{eq:entr}
\end{align} 
where we can now determine the entropy of each model parameter having 
\begin{align} 
\E_{q}\cor*{\lnp{q\pp*{\W}}} 
\eqeq \frac{\rm M C}{2}\lnp{2\pi e} + \frac{\rm M}{2}\ln\mid\Sigma_{\W}\mid
\end{align} 

\begin{align} 
&\E_{q}\cor*{\lnp{q\pp*{\alp}}} = \summ\Big(a_{\af} + \lnp{\Gamma\pp*{a_{\af}}} \nonumber\\
\eqline  - \pp*{1-a_{\af}} \psi\pp*{a_{\af}} - \lnp{b_{\af}}\Big)
\end{align} 

\begin{align} 
&\E_{q}\cor*{\lnp{q\pp*{\tau}}} \nonumber\\
\eqeq a_{\tau} + \lnp{\Gamma\pp*{\alpha_0^{\tau}}} - \pp*{1-\alpha_0^{\tau}} \pp*{\psi\pp*{\alpha_0^{\tau}} - \lnp{\beta_0^{\tau}}}
\end{align} 

\begin{align}
\E_q\cor*{\lnp{q\pp*{\bi}}} \eqeq \frac{\rm C}{2} \lnp{2\pi e} + \frac{1}{2} \ln\mid\Sigma_{\bi}\mid.
\end{align}

\subsection{Complete lower bound}
Finally, if we combine both terms, equation \eqref{eq:ElogpXTheta} and \eqref{eq:entr}, we get that the complete lower bound is 
\begin{align} 
&LB =
 - \pp*{\frac{C}{2}+\alpha_0-1}\summ\pp*{\lnp{b_{\af}}}   \nonumber \\
\eqline- \pp*{\alpha_0^{\tau}-1}\lnp{b_{\tau}} - \frac{1}{2}\ang{\bi\biT} - \beta_0^{\tau}\frac{a_{\tau}}{b_{\tau}}+\frac{C}{2}\pp*{\E_{q}\cor*{\lnp{\tau}}} \nonumber\\
\eqline - \frac{\ang{\tau}}{2} \sumn\sumc \Big(\frac{1}{2} \ang{\bic \bic} + \Ync\ang{\WcT} \phirff(\XnT) - \Ync \ang{\bic}\nonumber \\
\eqline -\frac{1}{2}\ang{\Wc\WcT} \phirff(\XnT)\phirff(\Xn)- \phirff(\Xn)\ang{\Wc} \ang{\bic}  \Big)\nonumber \\
\eqline - \frac{C}{2}\ln\mid\Sigma_{\W}\mid - \frac{1}{2} \ln\mid\Sigma_{\bi}\mid + \summ\lnp{b_{\af}}
 + \lnp{b_{\tau}} +\const, \label{eq:LqModelall}
\end{align}
where we can use Equation \eqref{eq:bTauBias} to simplify the lower bound
\begin{align} 
&LB =
 - \pp*{\frac{C}{2}+\alpha_0-1}\summ\pp*{\lnp{b_{\af}}}   \nonumber \\
\eqline- \pp*{\frac{C}{2} + \alpha_0^{\tau}-1}\lnp{b_{\tau}} - \frac{1}{2}\ang{\bi\biT} \nonumber \\
\eqline - \frac{C}{2}\ln\mid\Sigma_{\W}\mid - \frac{1}{2} \ln\mid\Sigma_{\bi}\mid + \summ\lnp{b_{\af}}
 + \lnp{b_{\tau}} +\const \label{eq:LqModel}
\end{align}
\subsection{Lower bound dependent on \texorpdfstring{$\Phirff$}{K}}
To maximise the lower bound to obtain the optimum $\gamma$ value we need the terms dependent on $\Phirff$ from equation \eqref{eq:LqModelall}, obtaining
\begin{align}
    LB \eqeq  \ang{\tau} \sumn\sumc \Big( \Ync\ang{\WcT} \phirff(\Xn)^{\top} - \phirff(\Xn)\ang{\Wc} \ang{\bic} \nonumber \\
    \eqline -\frac{1}{2}\ang{\Wc\WcT} \phirff(\Xn)^{\top}\phirff(\Xn) \Big) \label{eq:LB_K}
\end{align}

\section*{Acknowledgments}
The authors acknowledge support from the Spanish State Research Agency (MCIN/AEI/10.13039/5011000110) through project PID2020-115363RB-I00.

\bibliographystyle{elsarticle-num} 
\bibliography{bibliography}





\end{document}